Title Page:

# Evaluating LLMs in Medicine: A Call for Rigor, Transparency

Mahmoud Alwakeel MD[1], Aditya Nagori PhD[2], Vijay Krishnamoorthy MD, PhD[2], Rishikesan Kamaleswaran PhD[2]

[1] Department of Medicine, Duke University Hospital System, Durham, North Carolina, USA

[2] Department of Anesthesiology, Duke University Hospital System, Durham, North Carolina, USA

**Word Counts:** 1381

**Corresponding author:**

**Mahmoud Alwakeel, MD**

**Mailing address:** 2301 Erwin Rd, Durham, North Carolina, Zip Code 27705, United States

**Email:** mia21@duke.edu

**Abstract**

**Objectives:**

To evaluate the current limitations of large language models (LLMs) in medical question answering, focusing on the quality of datasets used for their evaluation.

**Materials and Methods:**

Widely-used benchmark datasets, including MedQA, MedMCQA, PubMedQA, and MMLU, were reviewed for their rigor, transparency, and relevance to clinical scenarios. Alternatives, such as challenge questions in medical journals, were also analyzed to identify their potential as unbiased evaluation tools.

**Results:**

Most existing datasets lack clinical realism, transparency, and robust validation processes. Publicly available challenge questions offer some benefits but are limited by their small size, narrow scope, and exposure to LLM training. These gaps highlight the need for secure, comprehensive, and representative datasets.

**Conclusion:**

A standardized framework is critical for evaluating LLMs in medicine. Collaborative efforts among institutions and policymakers are needed to ensure datasets and methodologies are rigorous, unbiased, and reflective of clinical complexities.

**Manuscript:**

**Introduction:**

Large language models (LLMs) are gaining attention in the medical field, where major tech companies promote their ability to accurately answer questions from medical exams.[1–3] These abilities are often highlighted as transformative for medical education and clinical practice. However, a recent randomized clinical trial (RCT) published in JAMA Network Open found that using ChatGPT-4 did not necessarily enhance physicians' diagnostic reasoning beyond conventional resources.[4] This finding highlights the need for a critical assessment of these models' real-world effectiveness and an exploration of the reasons behind discrepancies between their high performance in controlled settings and actual clinical applications.

In our opinion, a major concern lies in the databases used to evaluate and fine-tune LLM performance, as they are crucial for assessing reliability but often lack rigorous scrutiny. Examining these datasets is essential to identify potential sources of bias that could mislead evaluations and contribute to overpromising the capabilities of these models. Without such critical examination, it becomes difficult to trust the models in real-world medical scenarios where accuracy is paramount. Poorly validated LLMs could lead to misdiagnoses or inappropriate medical advice, and at best, they may impose unnecessary financial burdens on healthcare systems and patients, as highlighted by the findings of the recent RCT.[4]

This perspective examines widely used benchmark datasets such as Medical Question Answering (MedQA)[5], Medical Multiple-Choice Question Answering (MedMCQA)[6], PubMed Question Answering (PubMedQA)[7], and Massive Multitask Language Understanding (MMLU).[8] These datasets, frequently cited in academic literature and promoted by developers of large-scale LLMs platforms, are often presented as benchmarks for performance evaluation.[1,2] However, their widespread adoption necessitates a deeper evaluation of their integrity, rigor, and relevance to real-world clinical applications.

**Assessing the Fidelity of Benchmark Datasets: Composition and Realism:**

- **MedQA**[5]**:** This dataset contains 12,723 questions designed in the style of the United States Medical Licensing Examination (USMLE) questions, with only 300 sourced directly from tutorial materials provided by the exam company. The remaining questions were sourced from commercial websites, but no detailed information is provided regarding the authorship, the scientific rigor in their creation, or the rationale behind the answers. Additionally, there is a lack of transparency about the quality control or validation processes employed. Beyond the 300 tutorial-based questions, the dataset developers developed a retrieval system to extract reasoning for correct answers from textbooks. However, this system was validated on only 100 randomly selected questions and failed to provide complete evidence for 76% of them, according to the authors. Notably, one example included a medically incorrect answer that was published in the authors' paper without being recognized as incorrect by the creators. Despite these limitations, this dataset has been widely adopted as a benchmark for evaluating the performance of LLMs on medical questions in academic literature.[1–3]

- **MedMCQA**[6]**:** This dataset comprises 193,155 questions that include review materials and questions that are purportedly from actual exams for entry into India's postgraduate medical courses, such as All India Institute of Medical Sciences Postgraduate (AIIMS PG) and National Eligibility cum Entrance Test Postgraduate (NEET PG). However, it remains unclear whether these are legitimately sourced exam questions or how such exam questions are publicly accessible. The questions are typically concise, averaging 12 tokens in length, which limits their ability to reflect the complexity of real clinical scenarios. Additionally, the dataset shares similar validation and accuracy issues with the MedQA dataset, lacking thorough details on the processes of question development, answer formulation, and explanation verification.

- **PubMedQA**[7]**:** Developed from PubMed abstracts, this dataset is designed to answer straightforward questions based on single abstracts rather than synthesizing evidence from multiple sources. Only 0.4% of the dataset has been annotated by humans, with the remaining 22% unannotated and 77% annotated by artificial intelligence, without any quality assurance.

- **MMLU (Massive Multitask Language Understanding)**[8]**:** Comprising 15,908 questions covering 57 different subjects, including medical and non-medical topics, this dataset similarly lacks detailed information on the validity and accuracy required to reliably reflect medical exams or real-life situations.

**Pioneering a New Era: RAG in Synthetic Medical Question Development**

The emergence of Retrieval-Augmented Generation (RAG) represents a promising new technique in developing synthetic medical questions, potentially ushering in a transformative era for training and evaluating LLMs.[9] By combining information retrieval from medical resources with generative capabilities, RAG offers an opportunity to create clinically relevant questions. However, this innovation brings challenges that developers must carefully address. Intellectual property (IP) and copyright laws pose significant hurdles when utilizing textbooks or proprietary materials, requiring thoughtful navigation to ensure ethical and legal compliance.[10]

Moreover, while medical textbooks often present diseases as distinct, isolated entities, clinical reality is far more complex. A common saying in practice highlights this divergence: "Patients don't follow textbooks." This is especially true for patients presenting to hospitals or intensive care units (ICUs), where they uncommonly come with a single problem but instead exhibit multiple, intersecting issues. These cases demand not only medical knowledge but, more critically, clinical experience—key element that LLMs inherently lack. While developing synthetic databases using RAG may be practical for foundational and straightforward questions, creating more clinically mature questions that reflect the complexities of real clinical practice still requires significant refinement and further development.

**The Need for Building Robust Frameworks for LLM Evaluation in Medicine**

Unlike textbook-based approaches, clinicians who design board exams and high-quality in-training questions rely on real clinical vignettes and their expertise to craft questions that reflect the complexity and unpredictability of real-world patient scenarios. This is crucial because real-life cases often involve multiple, intersecting problems rather than isolated conditions. For LLMs to be validated effectively, their testing datasets should ideally mirror this complexity. However, creating such robust datasets is a highly time-consuming process that demands collaboration among multiple teams. For instance, the creation of each American Board of Internal Medicine

(ABIM) certification exam subspecialty involves coordination between an Item-Writing Task Force, mentors, and an approval committee.[11] Even after these datasets are developed, maintaining their integrity and preventing leakage to the internet poses significant challenges. Without such precautions, future LLMs could be trained on these datasets, compromising their utility for unbiased validation and testing.

Recognizing these challenges, researchers have explored alternative approaches, such as using challenge questions published in medical journals.[12] While these questions often reflect real-life cases and are more representative of real-world scenarios, they present notable limitations. Chief among these is their public availability, which allows LLMs to encounter them during training, diminishing their utility as unbiased evaluation tools. Additionally, the small number of these questions limits their applicability across a broader range of medical specialties, reducing their generalizability.

A practical solution to these issues would involve forming a dedicated organization to oversee the development of validated and trusted testing environments for LLMs. This organization could facilitate the secure, local, and independent testing of models without compromising dataset integrity. By maintaining tight control over datasets, such an initiative could ensure that LLM evaluations remain unbiased and reflective of clinical realities. As the utility of LLMs in medicine remains in its early stages, it is imperative for institutions and policymakers to develop a standardized framework for evaluating the ability of LLMs to answer medical questions. Such a framework would foster trust and transparency in LLM evaluations by ensuring that testing methodologies are rigorous, consistent, and representative of the complexities of clinical practice. Furthermore, this framework could guide the creation of secure and diverse datasets tailored to assess LLM performance across various specialties while safeguarding against data leakage or misuse. By addressing these foundational challenges, the medical community can pave the way for the responsible and effective integration of LLMs into clinical education and practice.

**Conclusion**

   The rapid integration of LLMs in medical applications highlights two critical concerns: the accuracy and validity of the datasets used in their training and evaluation, as well as the potential biases these datasets may introduce. Ensuring rigorous validation of these datasets is paramount to developing models that are both reliable and safe for clinical use. Without such scrutiny, biases in the training and validation data can undermine the effectiveness of LLMs, leading to inconsistent or even harmful outcomes in real-world applications. Addressing these challenges requires collaboration among researchers, technology developers, and regulatory bodies to establish robust standards for dataset integrity and validation. By prioritizing these measures, the medical community can better leverage the transformative potential of LLMs while minimizing risks and fostering trust in their application.